\documentclass[10pt,twocolumn,letterpaper]{article}

\usepackage{cite}
\usepackage{amsfonts}
\usepackage{textcomp}
\usepackage{xcolor}
\usepackage{cvpr}
\usepackage{times}
\usepackage{epsfig}
\usepackage{amsmath}
\usepackage{amssymb}
\usepackage{graphicx}
\usepackage{multirow}
\usepackage{subfig}
\usepackage{makecell}
\usepackage{stfloats}
\usepackage{algorithm}
\usepackage{algorithmic}
\hypersetup{
  pagebackref=true,breaklinks=true,letterpaper=true,colorlinks,bookmarks=false
}

\cvprfinalcopy 

\def\cvprPaperID{****} 

\begin{document}

\title{Voice-Face Cross-modal Matching and Retrieval: A Benchmark}

\author{Chuyuan Xiong\thanks{The first two authors contributed equally.}\\
Renmin University of China\\
{\tt\small chuyuan@ruc.edu.cn}
\and
Deyuan Zhang\footnotemark[1]\\
Shenyang Aerospace University\\
{\tt\small dyzhang@sau.edu.cn}
\and
Tao Liu\thanks{Corresponding author: tliu@ruc.edu.cn}\\
Renmin University of China\\
{\tt\small tliu@ruc.edu.cn}
\and
Xiaoyong Du\\
Renmin University of China\\
{\tt\small duyong@ruc.edu.cn}
}

\maketitle
\begin{abstract}

Cross-modal associations between person’s voice and face can be learnt algorithmically, which can benefit a lot of applications. The problem can be defined as voice-face matching and retrieval tasks. Much research attention has been paid on these tasks recently. However, this research is still in the early stage. Test schemes based on random tuple mining tend to have low test confidence. Generalization ability of models can’t be evaluated by small scale datasets. Performance metrics on various tasks are scarce. A benchmark for this problem needs to be established. In this paper, first, a framework based on comprehensive studies is proposed for voice-face matching and retrieval. It achieves state-of-the-art performance with various performance metrics on different tasks and with high test confidence on large scale datasets, which can be taken as a baseline for the follow-up research. In this framework, a voice anchored L2-Norm constrained metric space is proposed, and cross-modal embeddings are learned with CNN-based networks and triplet loss in the metric space. The embedding learning process can be more effective and efficient with this strategy. Different network structures of the framework and the cross language transfer abilities of the model are also analyzed. Second, a voice-face dataset (with 1.15M face data and 0.29M audio data) from Chinese speakers is constructed, and a convenient and quality controllable dataset collection tool is developed. The dataset and source code of the paper will be published together with this paper.

\end{abstract}

\section{Introduction}

Studies in biology and neuroscience have shown that human’s appearances are associated with their voices \cite{smith2016matching,mavica2013matching, smith2016concordant}. Both the facial features and voice-controlling organs of individuals are affected by hormones and genetic information \cite{harry, thornhill1997developmental, wells2013perceptions, kamachi2003putting}. Human beings have the ability to recognize this association. For example, when hearing from a phone call, we can guess the gender, the approximate age of the person on the other end of the line. When watching an unvoiced TV show, we can imagine the approximate voice by observing the face movement of the protagonist. With the recent advances of deep learning, face recognition models \cite{wen2016discriminative,wu2018light,liu2017sphereface} and speaker recognition models \cite{li2017deep, wang2018speaker} have achieved extremely high precision. Can the associations between voices and faces be discovered algorithmically by machines? The research on this problem can benefit a lot of applications such as synchronizing video faces and talking voice, generating faces according to voices.

\begin{figure}[!tbp]
\centering
\fbox{
\includegraphics[width=1\linewidth]{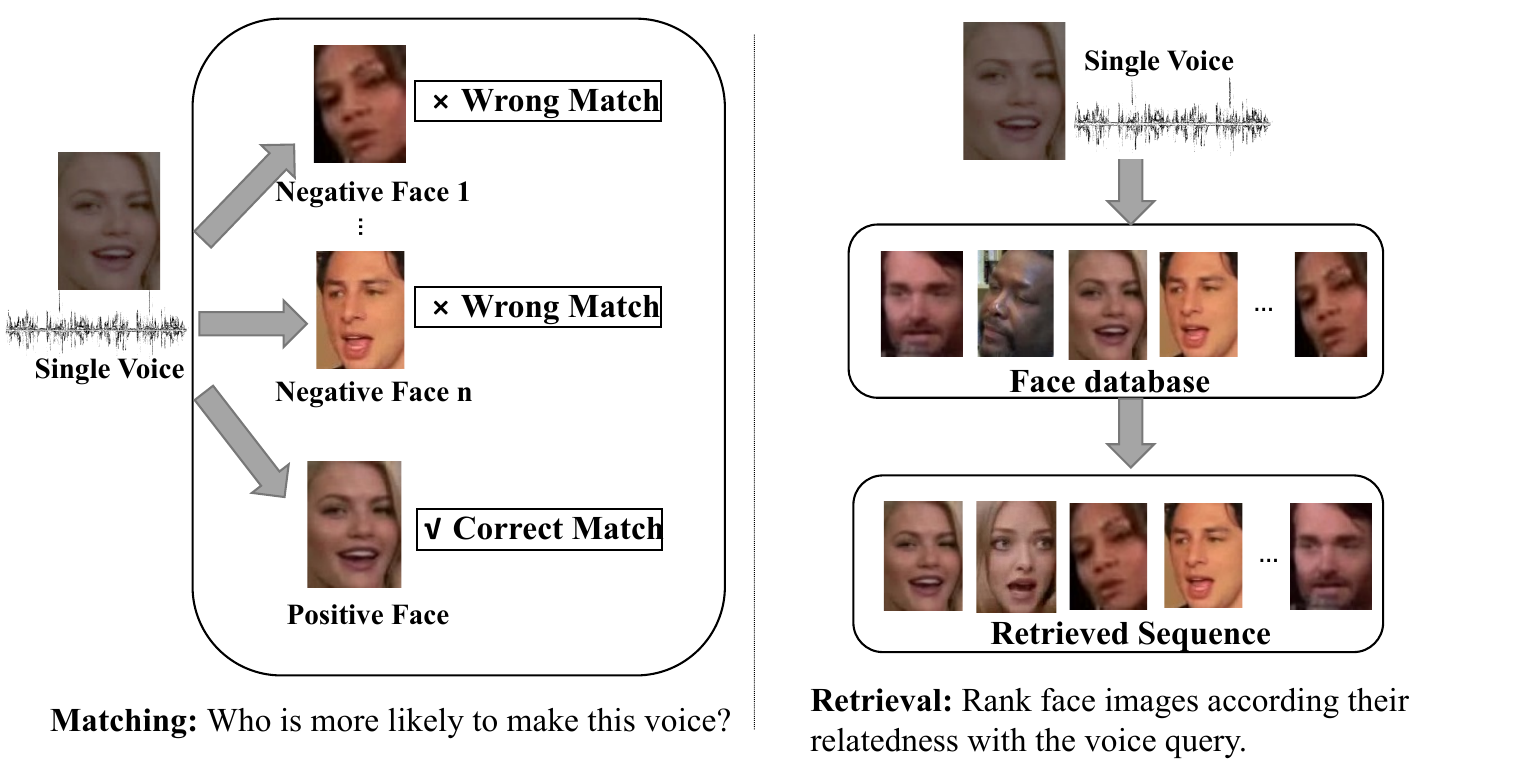}
}
   \caption{Voice-face matching and retrieval tasks.
}
\label{fig:mathchingandretrieval}
\end{figure}

In recent years, much research attention \cite{Nagrani2018Seeing, Wen2018Disjoint, Nagrani2018Learnable, horiguchi2018face, Kim2018On} has been paid on the voice-face cross-modal learning tasks, which have shown the feasibility of recognizing voice-face  associations. This problem is generally formulated as a voice-face matching task and a voice-face retrieval task as shown in Figure \ref{fig:mathchingandretrieval}. Given a set of voice audios and faces, voice-face matching is to tell which face makes the voice when machine hearing a voice audio. Voice-face retrieval is to present a sorted sequence of faces in the order of estimated match from a query of voice recording.

SVHF \cite{Nagrani2018Seeing} is the prior of voice-face cross-modal learning, which studies the performance of CNN-based deep network on this problem. The human’s baseline for voice-face matching task is also proposed in this paper. Both the “voice to face” and the “face to voice” matching tasks are studied in Pins \cite{Nagrani2018Learnable} and Horiguchi's work \cite{horiguchi2018face}, which exhibits similar performance on these two tasks. Curriculum learning schedule is introduced in Pins for hard negative mining. Various visualizations of embedding vectors are presented to show the learned audio-visual associations in Kim's work \cite{Kim2018On}.  DIMNet \cite{Wen2018Disjoint} learns the common representations for faces and voices by leveraging their relationship to some covariates such as gender and nationality. DIMNet obtains the accuracy of 84.12\% on the 1:2 matching which exceeds the human level.

The research on this problem is still in the early stage. Datasets used by previous research are always very small, which can’t evaluate the generalization ability of models sufficiently. Traditional test schemes based on random tuple mining tend to have low confidence. A benchmark for this problem needs to be established. This paper presents a voice-face cross-modal matching and retrieval framework, a dataset from Chinese speakers and a data collection tool. In the framework, cross-modal embeddings are learned with CNN-based networks and triplet loss in a voice anchored metric space with L2-Norm constraint. An identity based example sampling method is adopted to improve the model efficiency. The proposed framework achieves state-of-the-art performance on multiple tasks. For example, the result of 1:2 matching tested on 10 million triplets (thousands of people) achieved 84.48\%, which is also higher than DIMNet tested on 189 people. We have evaluated the various modules of the CNN-based framework and provided our recommendations. In addition, matching and retrieval based on the average of multiple voices and multiple faces are also attempted, which can further improve the performance. This task is a simplest way for analyzing video data. Large-scale datasets are used in this problem to ensure the generalization ability required in real application. The cross language transfer capability of the model is studied on the  voice-face dataset of Chinese speakers we constructed. A series of performance metrics are presented on these tasks by extensive experiments. The source code of the paper and the dataset collection tool will be published along with the paper.

\section{Voice-Face Matching and Retrieval (VFMR)}

\subsection{Dataset}

The dataset used by previous research is Vox-VGG-1 \cite{parkhi2015deep, Nagrani2017VoxCeleb, horiguchi2018face} with 1251 English speakers, which is the intersection of the VoxCeleb1 \cite{Nagrani2017VoxCeleb} dataset and VggFace1 \cite{parkhi2015deep} dataset. Vox-VGG-1 is also the intersection of the VoxCeleb1 dataset \cite{Nagrani2017VoxCeleb} and the MS-Celeb-1M dataset \cite{guo2016ms}. The generalization abilities of cross-modal learning models need to be evaluated on more data. In this paper, the Vox-VGG-2 \cite{cao2018vggface2,Chung2018VoxCeleb2} dataset formed by the intersection of the VoxCeleb2 \cite{Chung2018VoxCeleb2} dataset and the VGGFace2 \cite{cao2018vggface2} dataset is used, which includes 5994 English speakers. The Chinese\_VF dataset constructed by a data collection tool described in Section \ref{section22} is also used, which include 500 Chinese speakers. The statistic of Vox-VGG-1, Vox-VGG-2 and Chinese\_VF is shown in Table \ref{table:statistics}.
\begin{table}[!htbp]
\centering
\scalebox{0.65}{
    \begin{tabular}{cccccc}
    \hline
    Dataset & Gender& \#Identities & \#Utterances & \#Images (After MTCNN \cite{Zhang2016Joint}) \\
    \hline
    \multirow{3}{*}{Vox-VGG-2 \cite{Chung2018VoxCeleb2,cao2018vggface2},}
    & Male & 3,682 & 775,260 & 1,268,937 \\
    & Female & 2,312 & 316,749 &  636,079 \\
    & Total & 5,994 & 1,092,009 & 1,905,016 \\   
    \hline
    \multirow{3}{*}{Vox-VGG-1 \cite{Nagrani2017VoxCeleb,parkhi2015deep}}
    & Male & 690 & 103,295 & 378,152 \\
    & Female & 561 & 50,221 & 195,131 \\
    & Total & 1,251 & 153,516 & 573,283 \\  
    \hline
    \multirow{3}{*}{Chinese\_VF}
    & Male & 242 & 178,044 & 673,737 \\
    & Female & 258 & 118,225 & 476,973 \\
    & Total & 500 & 296,269 & 1,150,710 \\  
    \hline
    \end{tabular}
}
\caption{The statistics of voice-face cross-modal datasets.}
\label{table:statistics}
\end{table}

\subsection{Dataset Collection Pipeline}
\label{section22}

\begin{figure*}[!tbp]
\centering
\includegraphics[width=.75\linewidth]{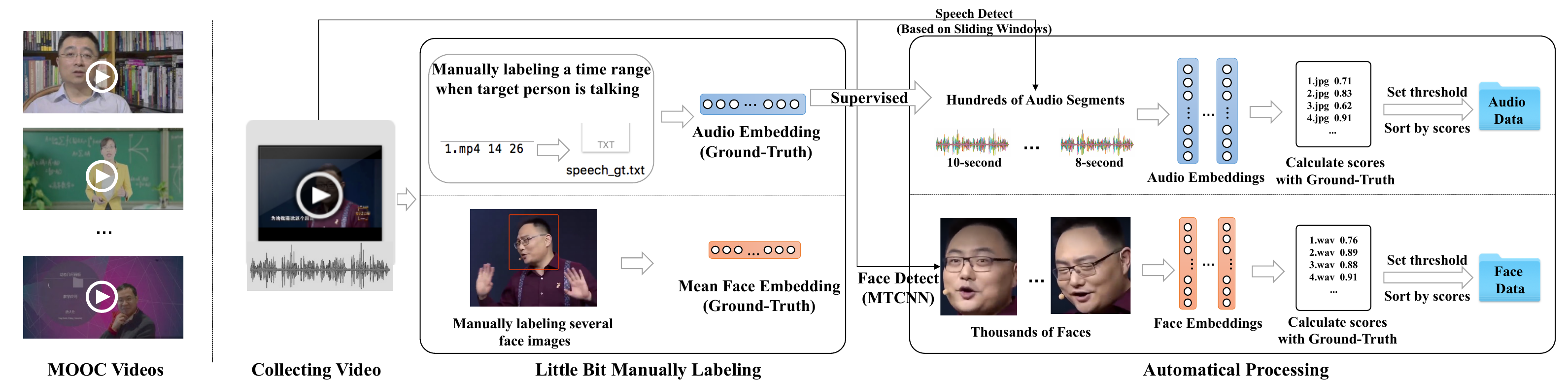}
\caption{Overview of the proposed dataset collection tool.}
\label{fig:collection}
\end{figure*}

Collecting datasets from speakers with different languages is important for analyzing the characteristics of cross language cross-modal learning. There are still no currently available voice-face datasets from Chinese speakers. Though some speech recognition datasets can be used for speaker recognition, there are no corresponding face images or videos available. Furthermore, the public available speech recognition datasets from Chinese speakers are also very rare and only contain a few speakers. A semi-automatic tool is developed for collecting voice-face data from type-specific videos. Collecting data from any kind of video will lead to very much noise and need tedious human labeling effort, and more complex techniques such as SyncNet\cite{chung2018learning} are needed to determine the active speaker. The type of videos for this tool is restricted to videos where only one person is talking in each video, which can greatly simplify the data collection procedure. The videos of MOOCs can meet this demand.  As shown in Figure \ref{fig:collection}, the data collection tool consists of five steps: 1) User need to capture several high quality face screenshots for a target speaker. 2) User needs to mark a time period in the video when the speaker is active. The audio corresponding to the marked time period is extracted, and is taken as a ground truth audio. 3) Face detection is conducted using MTCNN \cite{guo2016ms} and audio detection is conducted based on algorithms \ref{algo}. 4) The scores of all collected faces and speech segments are computed based on their similarities with the ground-truth. 5) By setting up two thresholds, high quality data are retained.

\begin{algorithm}[h]
    \caption{Speech Detection Algorithm}
    \label{algo}
    \begin{algorithmic}[1]
    \STATE
    \textbf{Input:}
      Audio-track from video: $v$;
      Ground-truth: $gt$;  
      Detection threshold: $t$;  \
    \STATE 
    \textbf{Initialize:}
      Window: $w$;
      Min window size: $s_{min}$;
      Max window size: $s_{max}$;
      Step window size: $s_{step}$;\
    \STATE $w_{start}=0, w_{end}=s_{min}$\
    \WHILE{$w_{end}<v_{length}$}
    \IF{$score(w, gt)>t$ and $score([w_{start}, w_{end}+s], gt) > score(w, gt)$}
    \STATE $[w_{start}, w_{end}]=[w_{start}, w_{end}+s_{step}]$ \
    \ELSE
    \STATE Extract segment $w$, record.
    \STATE $[w_{start}, w_{end}]=[w_{end}, w_{end}+s_{min}]$
    \ENDIF
    \ENDWHILE
    \end{algorithmic}
\end{algorithm}

\subsection{Learning Voice Anchored Embedding in L2-constrained Metric Space}

\begin{figure*}[!ht]
\centering
\includegraphics[width=.8\linewidth]{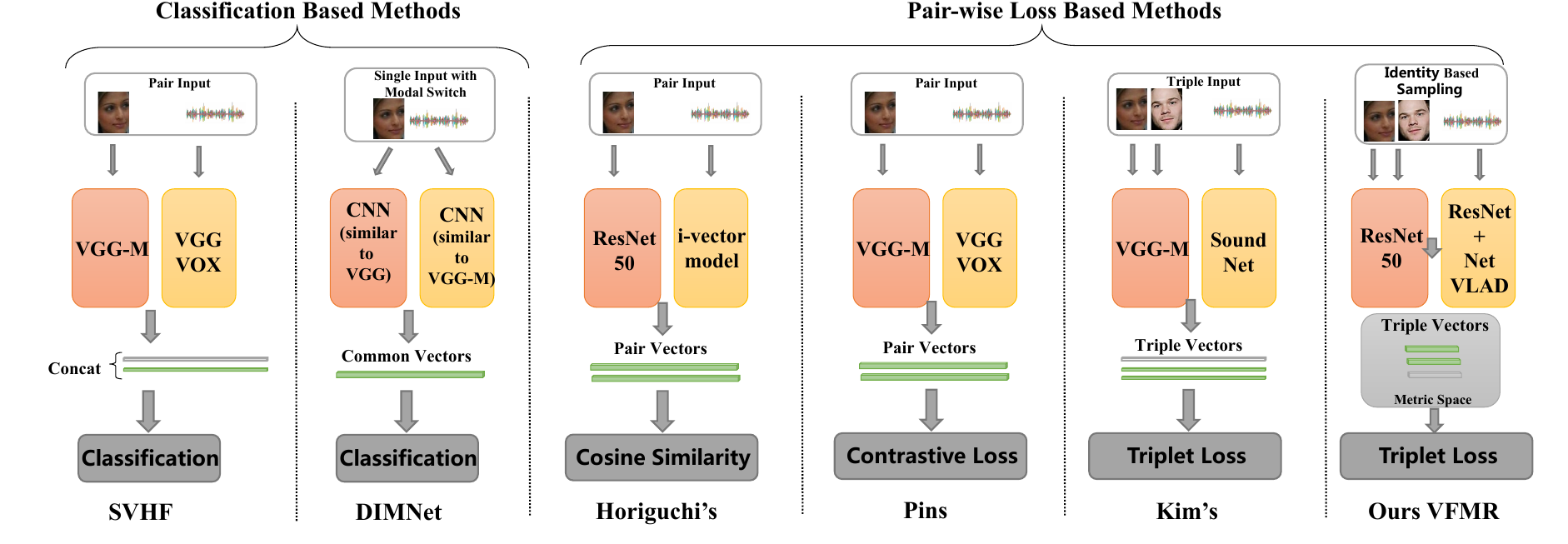}
\caption{Comparisons of VFMR with other existing methods.}
\label{fig:comparison}
\end{figure*}

The representation learning of voices and faces is the basis for voice-face matching and retrieval. The existing embedding learning methods for cross-modal learning can be classified as classification based methods and pair-wise loss based methods as shown in Figure \ref{fig:comparison}. CNNs-based networks are normally used to embed the voices and faces to feature vectors. In \cite{Nagrani2018Seeing}, feature embeddings from multi-stream CNN architecture are concatenated and then fed into multiple softmax classifiers for 1:n matching task. In \cite{Wen2018Disjoint}, controlled by a modality switch, voice or face embedding networks are selected to generate common features for either face or voice, and the learning is then supervised by a multi-task classification network. For pair-wise loss based methods, a pair or a triplet of vectors are embedded by voice and face network, and Contrastive Loss \cite{contrastive} or Triplet Loss \cite{triplet} is used to supervise the learning of embeddings. Pair-wise loss based methods are aimed at making the embeddings of positive pairs closer and the embeddings of negative pairs farther. Classification based methods are aimed at separating the embeddings of different classes. Compared with classification based methods, pair-wise loss based methods are better at distinguishing hard examples due to the characteristics of these methods. Adding additional supervised information for the classification based methods such as gender information are also targeted to constrain the example space. In this paper, a novel voice anchored embedding learning method in l2-constrained metric space is proposed, which belongs to the pair-wise loss based methods. The proposed method can improve the performance of traditional methods greatly. 

\begin{figure*}[!htbp]
\centering
\includegraphics[width=0.65\textwidth]{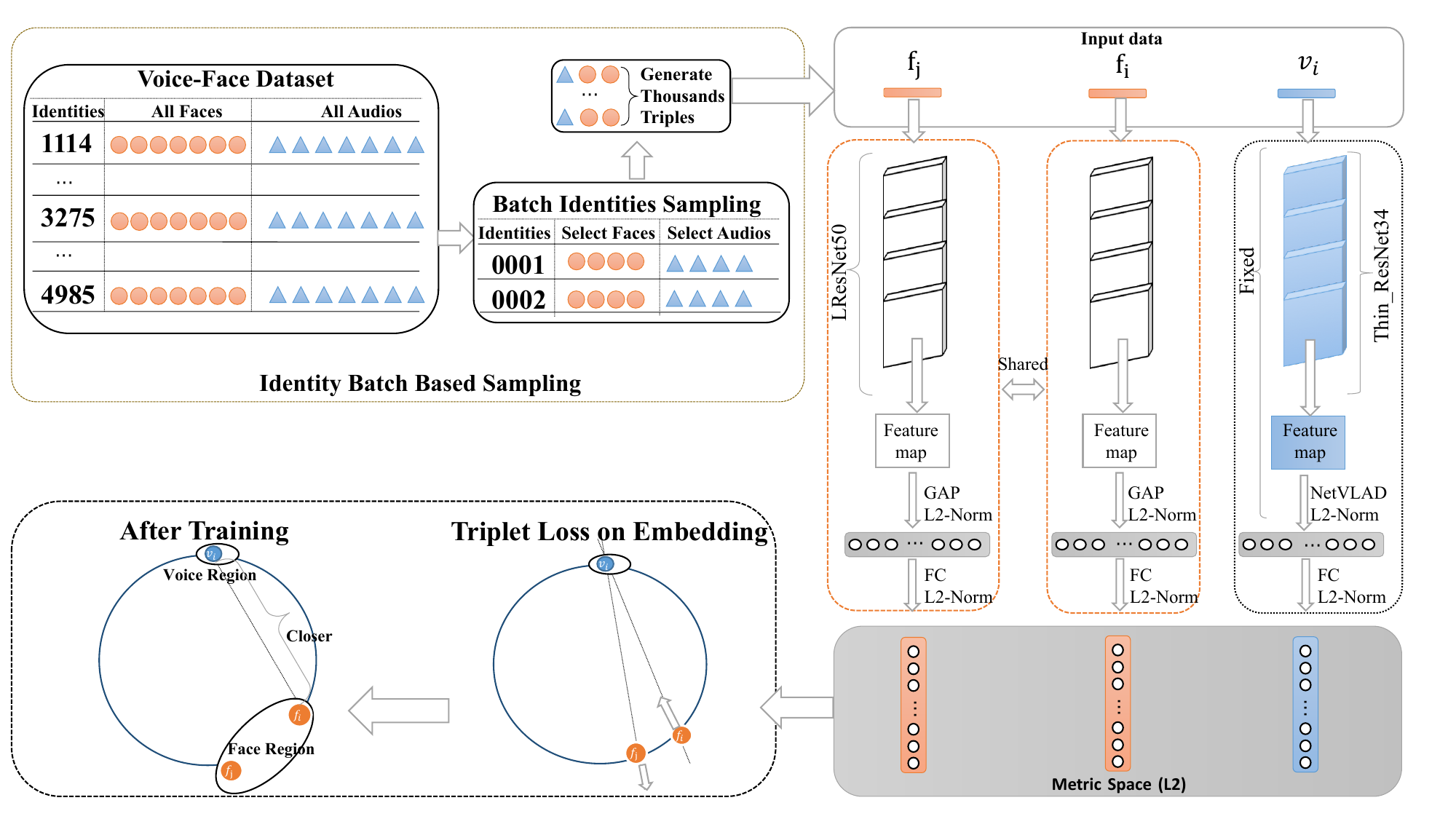}
\caption{Overview of Voice-face Embedding Learning Framework, which consists of three components. 1) Identity based data sampling method. 2) Face feature extraction based on LResNet50 and voice feature extraction based on Thin\_ResNet34 with NetVLAD. 3) Using Triplet Loss upon L2 constrained metric space. After training, face embedding and voice embedding form their own regions and the distance between positive samples tends to be closer.}
\label{fig:shortpipeline}
\end{figure*}

\begin{table}[!hbp]
\centering
\scalebox{0.7}{
    \begin{tabular}{c|c}
    \hline
    Thin\_ResNet34 & LResNet50\\
    \hline
    $ \rm{conv2d, 7\times 7, \quad 64}$ & \multirow{2}{*}{$ \rm{conv2d, 3\times 3, \quad 64}$} \\
    
    $ \rm{maxpool, 2\times 2}$ &\\
    \hline
    $\begin{pmatrix} 
      \rm{conv}, 1\times 1, \, 48 \\
      \rm{conv}, 3\times 3, \, 48 \\
      \rm{conv}, 1\times 1, \, 96 \\
      \end{pmatrix} \times 2$
       & $\begin{pmatrix} 
      \rm{conv}, 1\times 1, \, 64 \\
      \rm{conv}, 3\times 3, \, 64 \\
      \rm{conv}, 1\times 1, 256 \\
      \end{pmatrix} \times 3 $\\

    $\begin{pmatrix} 
     \rm{conv}, 1\times 1, \, 96 \\
     \rm{conv}, 3\times 3, \, 96 \\
     \rm{conv}, 1\times 1, 128 \\
      \end{pmatrix} \times 3 $
       & $\rm \begin{pmatrix} 
     \rm{conv}, 1\times 1, 128 \\
     \rm{conv}, 3\times 3, 128 \\
     \rm{conv}, 1\times 1, 512 \\
      \end{pmatrix} \times 4$\\

    $ \rm \begin{pmatrix} 
      \rm{conv}, 1\times 1, 128 \\
      \rm{conv}, 3\times 3, 128 \\
      \rm{conv}, 1\times 1, 256 \\
      \end{pmatrix} \times 3$
       & $\rm \begin{pmatrix} 
      \rm{conv}, 1\times 1, 256 \\
      \rm{conv}, 3\times 3, 256 \\
      \rm{conv}, 1\times 1, 1024 \\
      \end{pmatrix} \times 6$\\

    $ \rm \begin{pmatrix} 
      \rm{conv}, 1\times 1, 256 \\
      \rm{conv}, 3\times 3, 256 \\
      \rm{conv}, 1\times 1, 512 \\
      \end{pmatrix} \times 3$
       & $\rm \begin{pmatrix} 
      \rm{conv}, 1\times 1, 512 \\
      \rm{conv}, 3\times 3, 512 \\
      \rm{conv}, 1\times 1, 2048 \\
      \end{pmatrix} \times 3$\\
    \hline
    $\rm{maxpool, 3\times 1}$ & $\rm{FC2, 512}$\\
    \hline
    $\rm conv2d, 7\times 1, 512$ & $\rm BN+L2\_norm$\\
    \hline
    \end{tabular}
}
\caption{Structures of Thin\_ResNet34 and LResNet50.}

\label{table:backbone1}
\end{table}

\textbf{Network Structure.} LResNet50 \cite{Deng2018ArcFace} and Thin ResNet34 \cite{Xie2019Utterance} with NetVLAD \cite{Arandjelovic2017NetVLAD} are well-performed networks in face recognition and speaker recognition task respectively. These two networks as shown in Table \ref{table:backbone1} are used in this paper for face feature extraction and voice feature extraction respectively. The input of the embedding network is a triplet set. For a specific triplet $<v_i, f_i,f_j>$, $v_i$ and $f_i$ are from same identity, $v_i$ and $f_j$ are from different identities. The feature extraction functions for voice and face are defined as $Feature_v(v)$ and $Feature_f(f)$ respectively. A full connected layer is added to form the embedding vector as
$emb_v(v)=s\times \|(W_v\times Feature_v(v)+B_v)\|_2^2$, and $emb_f(f)=s\times \|(W_f\times  Feature_v(f)+B_f\|_2^2)$, we add l2 constraints and scale variables $s$ to the output of the network.

\begin{figure}[htbp]
  \centering
  \subfloat[Original]{
    \label{fig:subfig:a}
    \includegraphics[width=0.3\linewidth]{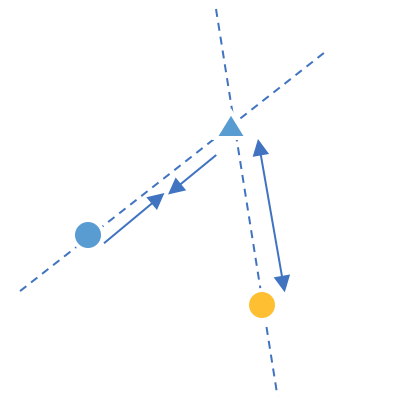} 
    }
  \subfloat[L2-Norm]{
    \label{fig:subfig:b}
    \includegraphics[width=0.3\linewidth]{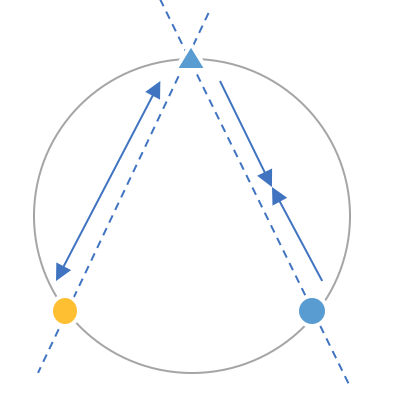}
    }
  \subfloat[Voice Anchored L2-Norm]{
    \label{fig:subfig:c}
    \includegraphics[width=0.3\linewidth]{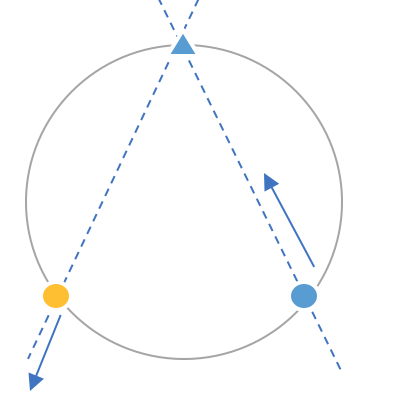}
    }
  \caption{ Visualization of three metric learning methods.}
  \label{fig:subfig} 
\end{figure}

\textbf{Triplet loss upon l2-constrained metric space.} Triplet loss is adopted in this paper for embedding learning. As illustrated in Figure \ref{fig:subfig:a}, embedding vectors from the same person in Euclidean space will be closer after long time training. Since there are billions of input triplets, it is difficult to obtaining satisfactory results by directly training in the huge Euclidean space. Two strategies are adopted in this paper to deal with this problem. First, L2 normalization is added to constrain the embedding vectors to be on a spherical space (Figure \ref{fig:subfig:b}). Second, voice anchored embedding learning is adopted. By freezing the pre-trained voice embedding network, feature vectors from voice are served as anchors and the goal of the model is to make the positives approach and keep negatives away (Figure \ref{fig:subfig:c}). Examples tend to be distinguished much better and faster in voice anchored embedding learning process together with the L2 constrained space. Suppose $d(x)$ indicates Euclidean distance, the loss function is defined as Eq. \ref{formula5}, where $m$ is a margin to control the distance between positive pairs and negative pairs. 

\begin{equation}
\scalebox{0.8}{
\begin{math}
\begin{aligned}
Loss = &\sum_{v_i,f_i,f_j, i\neq j}^{Batch}{}{\max{[d(emb_v(v_i), emb_f(f_i)) }}\\ &{{- d(emb_v(v_i), emb_f(f_j)) + m}, 0]}
\end{aligned}
\end{math}}
\label{formula5}
\end{equation}

\subsection{Using Embedding for Matching and Retrieval}

The embedding networks $emb_v(v)$ and $emb_f(f)$ can be used to extract embeddings from voice audios and face images. A similarity between a voice embedding and a face embedding is computed as an inner product of the two vectors. Voice-face matching and retrieval are conducted based on the computed similarities.

\textbf{Voice-Face Matching.} A general form for voice-face cross-modal matching is 1:n matching. Given an audio $v$ and $n$ identities' face images $F=<f_1, f_2, ..., f_n>$, 1:n voice-face matching refers to find the face who make this voice $v$. 1:2 matching is a special case of 1:n matching when $n=2$ and it is a primary task for voice-face matching.

\textbf{Voice-Face Retrieval.} Given a “query” voice, voice-face retrieval is to rank face images according to their relevance with the voice query. The retrieval task is a supplement to the matching task. The retrieval candidates always contain multiple face pictures for one person. The position-related information of all the positive faces is more effective for analyzing the model performance. Up to date, there are still very rare results on this problem published by the related works. The retrieval task is significant for many practical applications. For example, with the audio of a suspect, the police can retrieve the face of the suspect from a faces database based on the audio.

\textbf{Joint-voice or joint-face learning.} Instead of single audio segment or single face for one identity, multiple audios and faces or videos can provide more information. Joint-voice or joint-face matching and retrieval task are defined in this paper. The “joint voice” or “joint face” refers to combine the multiple audios or images by calculating the mean embeddings. This task can be used for video scenes.

\subsection{Identity Based Sampling for Training and Test Efficiency}
\label{Tfrom}

\textbf{Identity based sampling v.s. Random tuple mining.} The input triplets for the voice-face embedding network need to be mined from the datasets, the number of which is extremely large. Suppose $n$ identities are selected, the number of faces for per identity is $n_f$ , and the number of audios is $n_v$, then there will be a total of $n\times (n-1) \times n_v \times n_f^2$ triplets. Random tuples mining also named as “offline mining” will lead to training and test inefficiency. Identity based sampling named as “online mining” is adopted in this paper, which can greatly improve the training and testing efficiency. In the identity based sampling, a batch of identities is randomly selected first, and then certain number of face images and audios for each identity of the batch are sampled. Triplets are generated based on each batch of identities. Triplet Loss is susceptible to noise which means direction of network convergence is easy to change by few noise samples. Identity based training can effectively handle the disadvantage of Triplet Loss.

\textbf{Testing confidence coefficient.} It is important to guarantee the confidence of testing results produced by example sampling. It should be discovered that how many triplets need to be sampled to qualify the confidence of testing results. For random tuple mining method, suppose the test dataset contains $N$ identities, the rate of two identities A and B to be jointly selected to form a test tuple $t=<v_A, f_A, f_B>$ is $\frac{1}{N\times(N-1)}$. The expected frequency of two identities A and B to be jointly selected as test tuples in the complete test process is denoted as $K$. Suppose totally $n$ test tuples are mined. $K=\frac{n}{N\times(N-1)}$ can be derived. By analyzing the quantitative relation among the shift of $K$, $N$ and testing confidence, it can be derived that testing confidence is proportional to $N$ and $\ln{K}$. We can introduce the coefficient $T$ to simply measure the confidence. 
\begin{equation}
\scalebox{0.8}{
$T=N\times \ln{K}=N\times \ln{\frac{n}{N\times (N-1)}}$
}
\label{formulaT}
\end{equation}
For identity based sampling method, suppose the number of identities in a batch is $b$, and $r$ faces and $q$ voices are assigned for each identity, the number of tuples formed by A and B in the batch is $q\times r^2$. It can be derived $K=\frac{C_m^2}{N\times (N-1)}\times b \times q \times r^2$. Since the total tuples number $n$ equals $q\times r^2\times 2C_m^2$, $K$ is still valid in the case of identity based sampling.

\section{Experiments}
\subsection{Training Settings and Details}

\begin{table*}
\centering
\scalebox{0.7}{
\begin{tabular}{c|c|c|c|c|c|c}
\hline
\multirow{8}{*}{Settings} & Methods & Train (\#Identities) & Nationality & Face network & Voice network & Loss  \\
\cline{2-7}
& SVHF \cite{Nagrani2018Seeing} & 942 & US/UK & VGG-M & VGG-Vox & Softmax  \\
& Pins \cite{Nagrani2018Learnable} & 901 & US/UK & VGG-M & VGG-Vox & Contrastive  \\
& DIMNet \cite{Wen2018Disjoint} & 924 & US/UK & DIMNet-face & DIMNet-voice & Softmax  \\
& Kim's \cite{Kim2018On} & 1101 & US/UK & VGG-M & VGG-Vox & Triplet  \\
& Horiguchi's \cite{horiguchi2018face} & 862 & US/UK & ResNet50 & i-vector & Cosine  \\
& VFMR1 & 5994 & US/UK   & LResNet50 & Thin\_ResNet34+NetVLAD & Triplet\\
& VFMR2 & 1000 & US/UK   & LResNet50 & Thin\_ResNet34+NetVLAD & Triplet\\
& VFMR3 & 5994 & US/UK   & LResNet50 & Thin\_ResNet34+NetVLAD & Triplet\\
& VFMR4 & 400  & China & LResNet50 & Thin\_ResNet34+NetVLAD & Triplet\\
\hline
\hline
\multirow{8}{*}{1:2 Matching} & Methods &Test (\#Identities) & Nationality & $n$ triplets & Confidence $T$ & Value (ACC\%) \\
\cline{2-7}
& SVHF \cite{Nagrani2018Seeing} & 189 & US/UK & 10k & -239 & 81.0  \\
& Pins \cite{Nagrani2018Learnable} & 100 & US/UK & - & - & 78.5(AUC\%)  \\
& DIMNet-I \cite{Wen2018Disjoint} & 189 & US/UK & 678M & 992 & 83.45  \\
& DIMNet-IG \cite{Wen2018Disjoint} & 189 & US/UK & 678M & 992 & 84.12  \\
& Kim's \cite{Kim2018On} & 250 & US/UK & - & - & 78.2  \\
& Horiguchi's\cite{horiguchi2018face} & 216 & US/UK & 38B & 2188 & 78.10  \\
& VFMR1 & 1251 & US/UK   & 30.72M & 3725 & \textbf{84.48}\\
& VFMR2 & 189  & US/UK   & 3.72M & 842 & 83.55\\
& VFMR3 & 500  & China & 3.72M & 2502 & 71.52\\
& VFMR4 & 100  & China & 3.72M & 823 & 79.14\\

\hline
\hline
\multirow{8}{*}{Retrieval} & Methods & Test (\#Identities) & Nationality & \#query audios & Chance (mAP\%) & mAP \cite{christopher2008introduction} (\%) \\
\cline{2-7}
& DIMNet-IG \cite{Wen2018Disjoint} & 189 & US/UK & 21k & 1.07 & 4.42  \\
& Horiguchi's \cite{horiguchi2018face} & 216 & US/UK & 26k & 0.46 & 1.96  \\
& VFMR1 & 1251 & US/UK   & $1251\times 40=50k$ & 2.15 & \textbf{11.48}\\
& VFMR2 & 189  & US/UK   & $189\times 40=7k$ & 2.15 & 9.61\\
& VFMR3 & 500  & China   & $500\times 40=20k$ & 2.15 & 5.00\\
& VFMR4 & 100  & China & $100\times 40=4k$ & 2.15 & 7.06\\
\hline

  \end{tabular}
}
\caption{Comparison with other models on 1:2 matching task and retrieval task. $T$ is the confidence coefficient proposed in Section \ref{Tfrom}, which reflects the confidence level of the 1:2 matching. We recommend that the value of $T$ be given as a confidence level when performing an incomplete 1:2 matching test and higher confidence and generalization ability can be achieved with higher $T$.}
\label{table:compareall}
\end{table*}

\textbf{Training and test data split.} Four models with different training and testing splits are setup in the experiments as VFMR1, VFMR2, VFMR3and VFMR4. The training and test split for these models are depicted in Table \ref{table:compareall}.

\textbf{Data preprocessing.} Face images preprocessing includes face detection and the stretching scale of detection box. Face detection based on MTCNN \cite{Zhang2016Joint} is conducted on Vox-VGG-1 and Vox-VGG-2. For Chinese\_VF, face detection based on 1.3 scale MTCNN has been conducted during the dataset collection process. All face images are then rescaled to $112 \times 112 \times 3$ as the input for face embedding networks. Audio preprocessing consists of 512 point FFT, a short-time Fourier transform (STFT) for each frame and normalization. Though the voice embedding network based on Thin ResNet34 and NetVLAD can accept audio input with any length, training audios are uniformly cut to 2.5second for training efficiency and no clipping is conducted for testing audios. The input shape for a $k$-second audio clip is $257 \times (100\times k) \times 1$.

\textbf{Model settings.} The voice embedding network and face embedding network are pre-trained by VoxCeleb2 and VGGFace2 respectively. Margin $m$ for Triplet Loss is set to 1, and scale $s$ for L2 normalization is set to 128. Adam Optimizer is adopted in this paper. The total number of learning steps is 70k. The learning rate of FC layer for “step$<$20k”, “20k$<$step$<$40k”, “40k$<$step$<$60k” and “step$>$60k” is $10^{-3},10^{-4},10^{-5},10^{-6}$ respectively. The learning rate of face embedding network is fixed to $10^{-6}$.

\textbf{Settings for Different Task.} 1) 1:2 matching settings. The identity number $b$ of a batch is set to 4, the audios number $q$ and the faces number $r$ to be selected in a batch per identity is set to 4 and 8 respectively. The number of triplets in one batch is $(m\times q)\times r^2=3072$. A total of 10k steps (total of 30.72M triplets) are tested on VFMR1. As definited in Section \ref{Tfrom} and Eq. \ref{formulaT}, $K_{VSMR1}=19.65$ as the number of calculations between any two identities, and the $T_{VSMR1}=3725$ as the confidence coefficient. 1000 steps with 3M test triplets are tested for VSMR2 and other default tests. $K_{VSMR2}=86.46$, $T_{VSMR2}=842$. It should be noted that in order to balance the gender of the test triplets in 1:2 matching task, the gender ratio of test batch identities should be 3:1 or 1:3.  2) 1:n matching settings. The sample number of tuples will be much higher than triplets in 1:2 matching, so we performed this test directly on the 10k tuples, and the confidence level will be lower, but it can reflect the comparison result. 3) Retrieval settings. The face database comprises 500 pictures, which come from randomly selected 100 identities. 40 audio queries are constructed for each identity. 4) Joint matching and retrieval. Two variables $m\_f$ and $m\_v$ are respectively introduced to represent how many faces and audios are synthesized as a single face or audio. For joint-matching, various values of $m\_f$ and $m\_v$ are tested in experiments. For joint-retrieval, $m\_v$ is set to 20 and $m\_f$ is set to 5 directly.

\subsection{Performance Comparison}

Table \ref{table:compareall} presents the comparison of the proposed method with other related works on 1:2 matching and retrieval tasks. VFMR1 achieves state-of-the-art performance on both matching and retrieval tasks. To compare with existing approaches more directly, VFMR2 with almost the same training and testing data as related works is set up. VFMR2 outperforms the traditional pair-wise loss methods such as Kim’s \cite{Kim2018On} and Horiguchi’s \cite{horiguchi2018face} by more than 5 percent. The performance of pair-wise loss methods are always limited by small-scale training data. Identity based training and L2-norm based metric learning used by VFMR2 can greatly improve the training effectiveness of pair-wise loss based methods. Even though, training and testing on small datasets are not recommended. During the test process of VFMR1, when different subsets of 189 identities were selected instead of the total 1251 identities, we discovered the results obtained each time were very different. Small test dataset can’t reflect the generalization ability of the model. The confidence coefficient $T$ provided is for evaluating confidence level of model generalization ability. VFMR1 is recommended as a benchmark for voice-face 1:2 matching task, since it achieves the best results on large training and testing datasets.

\begin{figure}[!htbp]
\centering
\includegraphics[width=0.6\linewidth]{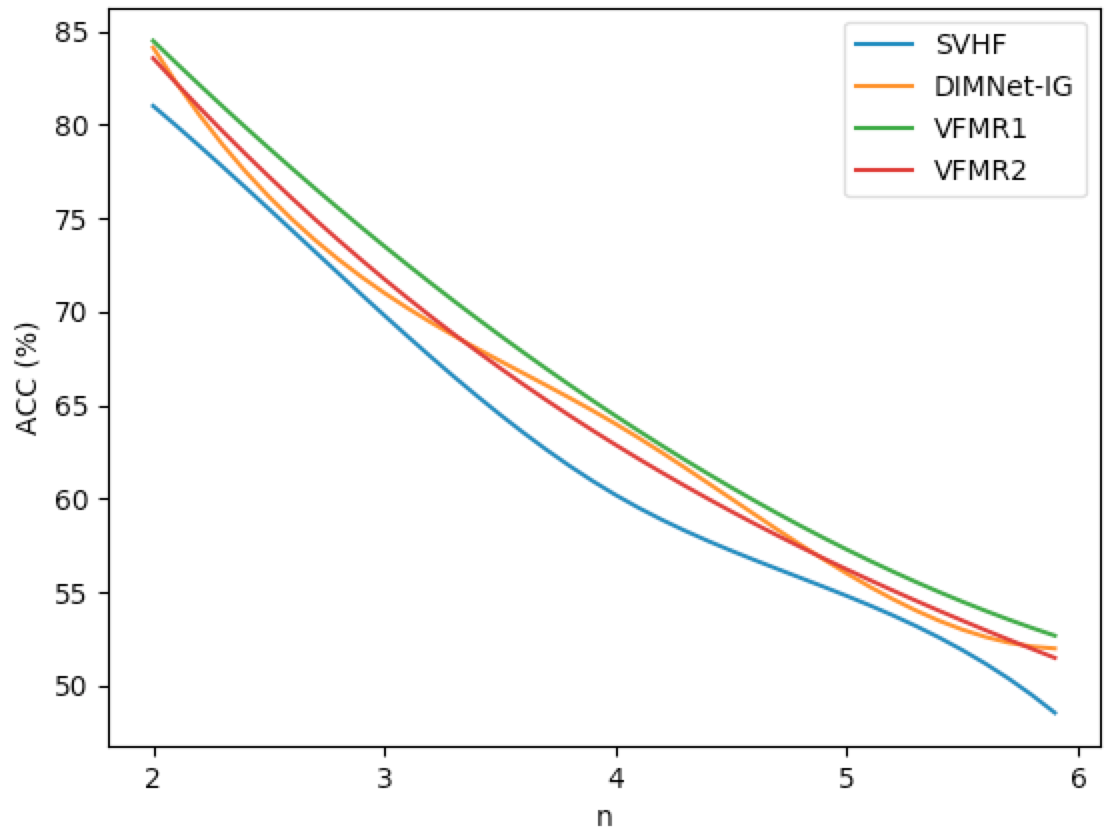}
\caption{Comparison with other available results by existing methods on 1:n matching task. Accuracy in all models decreases rapidly with n increases.}
\label{fig:midsubfig:1_n}
\end{figure}

\begin{figure}[!htbp]
\centering
\includegraphics[width=0.8\linewidth]{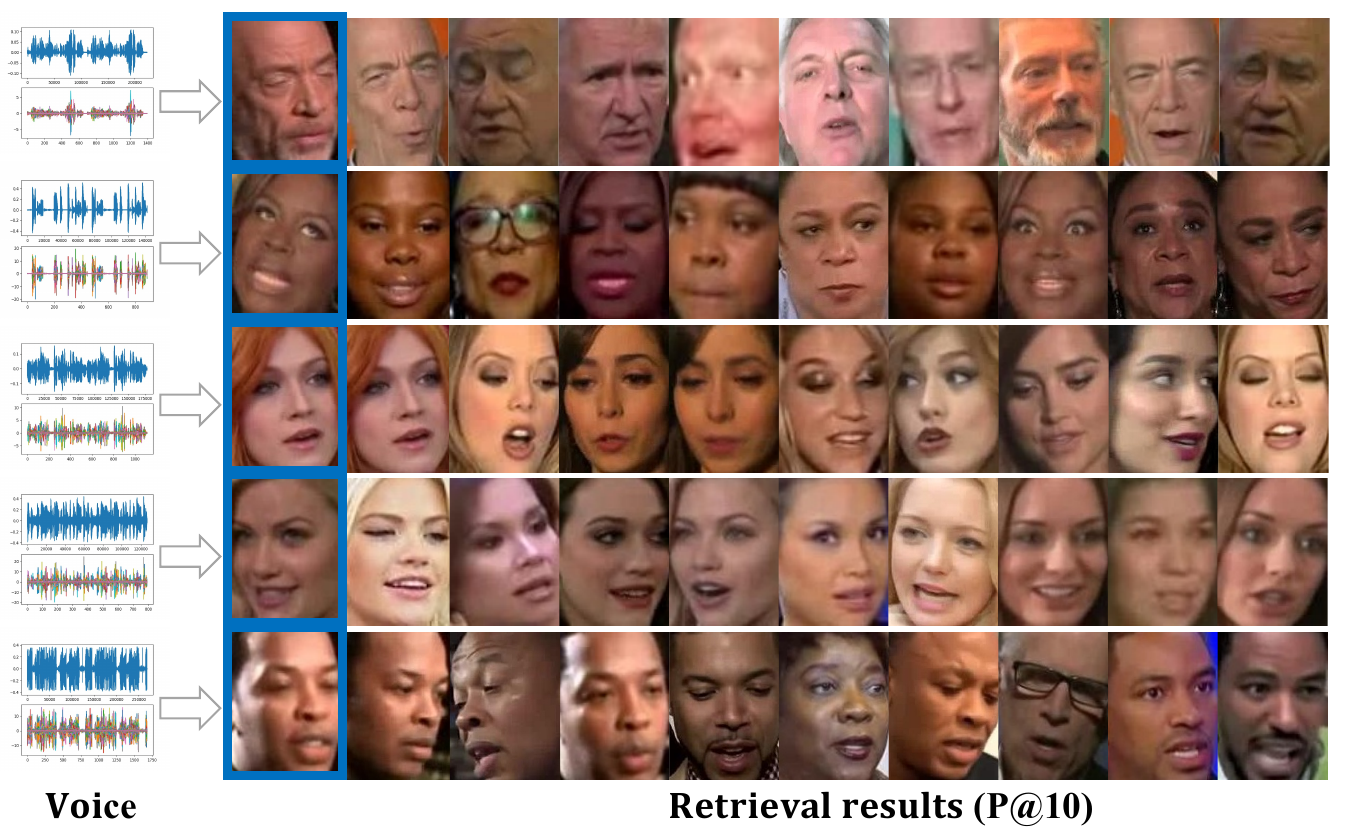}
\caption{Qualitative analysis of retrieval results produced by VFMR1.}
\label{fig:midsubfig:individual}
\end{figure}

Figure \ref{fig:midsubfig:1_n} shows the comparisons of VFMR with the currently available results from other methods on 1:n matching task. VFMR1 still performs best on this task. Figure \ref{fig:midsubfig:individual} visualizes some retrieval results of VFMR1 with $p@1=1$. From the illustration, we can see the top ranked faces are very similar, which shows the potential of voice-face cross-modal retrieval to be applied in real applications.

\subsection{Joint-matching and Joint-retrieval}

The results of joint-voice and joint-face matching are shown in Table \ref{table:1-nfaces}. The accuracy of 1:2 matching can be improved further by synthesizing voices and faces. Especially, synthesizing ten voices and ten faces can obtain the accuracy of 89.66\%, which is 5 percent higher than that of single voice and single voice matching. On VFMR2, the retrieval $mAP$ can be improved by 13 percent.

\begin{table}[!htbp]
\centering

\scalebox{0.6}{
    \begin{tabular}{cc|cc|cc||c|c|c}
    \hline
    \multicolumn{6}{c||}{Joint-Matching (VFMR1)} & \multicolumn{3}{c}{Joint-Retrieval} \\
    \hline
    $m_f$ & ACC & $m_v$ & ACC & $m_f+m_v$ & ACC & & Joint-Voice mAP & Joint-Face mAP\\
    \hline
    5 & 85.55& 5 & 85.13 & 1+1 & 84.48     & VFMR1 & 12.53 & 21.65\\
    10 & 84.56& 10 & 86.01 & 5+5 & 86.42   & VFMR2 & 12.57 & 22.83\\
    15 & 86.28& 15 & 85.49 & 10+10 & \textbf{89.66} & VFMR3 & 5.29 & 10.39 \\
    20 & 86.53& 20 & 86.16 & 20+20 & 89.54 & VFMR4 & 8.01 & 15.10  \\
    30 & 84.71& 30 & 85.63 & 30+30 & 89.28 & Random& 2.15 & 5.25\\
    \hline
    \end{tabular}
}
\caption{Performance on joint-face and joint-voice matching and retrieval task.}
\label{table:1-nfaces}
\end{table}

\subsection{Cross Language Transfer}

From the result of VFMR3, as shown in Table \ref{table:bigtable0}, we can see the model trained on dataset of English speakers obtains the accuracy of 71\% on Chinese identities. Compared to VFMR3, VFMR4 can improve the accuracy by nearly 8 percent, with only the fine-tuning effort on Chinese identities. Better results will be achieved after further expanding the data scale with the tools developed in this paper.

\begin{table}
\centering
\scalebox{0.7}{
\begin{tabular}{c|c|c|c|c|c|c}
\hline
 Model & 1:2  & 1:3 & 1:4 & Male & Female & mAP         \\
\hline
 VFMR3& 71.52 & 55.38 & 45.12 & 55.57 & 54.46 & 5.00 \\
 VFMR4& 79.14 & 67.10  & 57.80 & 64.49  & 64.04 & 7.06 \\ 
\hline
  \end{tabular}
}
\caption{Performance comparison of VFMR3 (Trained on Vox-VGG-2 and tested on Chinese\_VF) and VFMR4 (Five-fold cross validation on Chinese\_VF and the pre-training is the same as default settings).}
\label{table:bigtable0}
\end{table}

\begin{table}
\centering
\scalebox{0.7}{
\begin{tabular}{c|c|c|c|c|c|c}
\hline
 Model & 1:2  & 1:3 & 1:4 & Male & Female & mAP         \\
\hline
 VFMR1(UU)& 84.48 & 73.50 & 64.43 & 69.85 & 71.04 & 11.48 \\
VFMR1(SH)& 87.95 & 78.75 & 71.86 & 77.45  & 78.47 & 19.21 \\
VFMR2(UU)& 83.55 & 71.74 & 62.88 & 66.89 & 70.97 & 9.61 \\
VFMR2(SH)& 95.97 & 92.16 & 89.19 & 92.12 & 92.14 & 38.39 \\
\hline
  \end{tabular}
}
\caption{Performance comparison of Unseen Unheard(UU) test and Seen Heard(SH) test on various tasks.}
\label{table:bigtable1}
\end{table}

\subsection{Seen and Heard (SH) Test}

If test triplets are constructed from unused examples of seen and heard identities from training set, the testing is referred as seen and heard (SH) test. As shown in Table \ref{table:bigtable1}, 95.97\% accuracy can be obtained on seen and heard test of 1:2 matching task. SH test, which has much higher accuracy, presents its potential to be put into real applications. 

\subsection{Individual Test}

\begin{figure}[!htbp]
\centering
\includegraphics[width=1\linewidth]{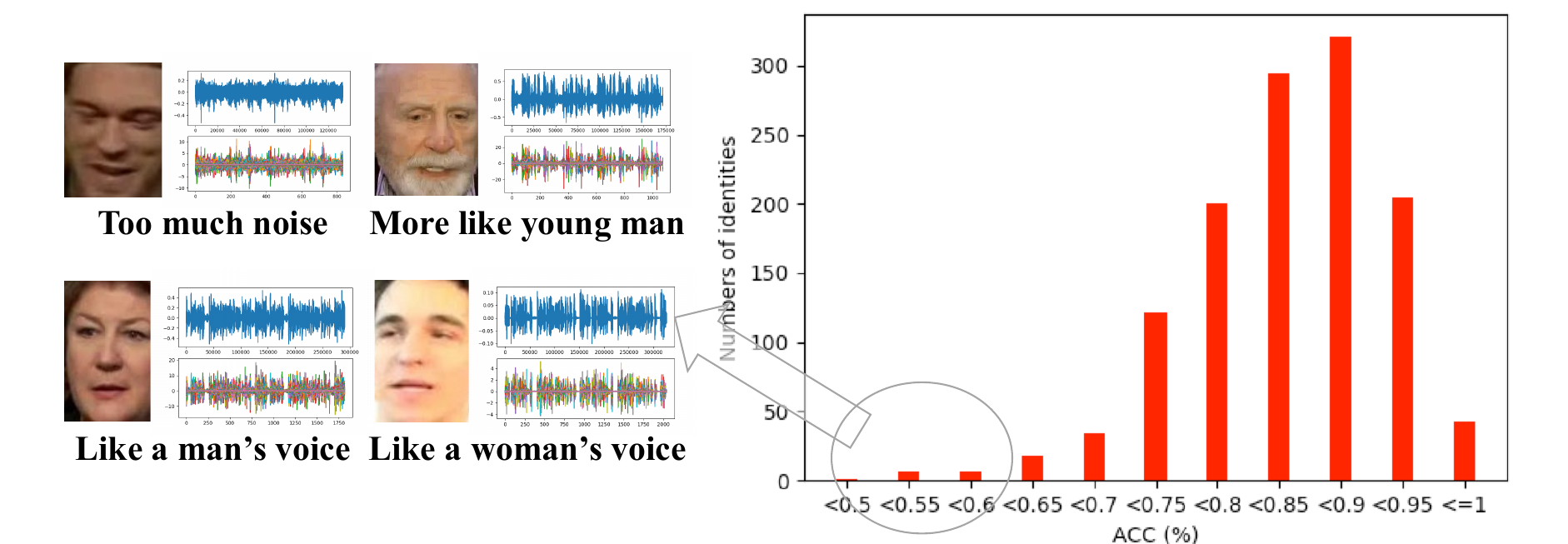}
\caption{Individual accuracy statistics of VFMR1 on 1:2 matching task. Very low accuracies occur on some abnormal examples.}
\label{fig:singleperson}
\end{figure}

Individual test on each identity is conducted to find the bottleneck of the model. Every identity of Vox-VGG-1 is taken as a target identity successively, and each of the remaining 1250 identities is in turn to form a pair with the target identity for 1:2 matching. The experiment is repeated ten times for each pair of identities. The average accuracy of the totally 12500 runs of every target identity is referred as the individual accuracy, the statistic of which is illustrated in the right part of Figure \ref{fig:singleperson}. The accuracy of most identities is above 75\%. There are only 15 identities whose accuracy is lower than 60\%. The left part of Figure \ref{fig:singleperson} illustrates the audios and faces of some of these identities. We discover most of the low accuracies occur on examples with noisy audios and some abnormal associations of faces and voices. 

\subsection{Comparisons of Different Submodule Settings}

\textbf{The effect of face detection.} We need to study whether to use face detection and what size of detection box should be used. As shown in Table \ref{table:default config} and Table \ref{table:bigtable2} without the use of face detection, too much noise will be introduced along with a few useful features and the performance of VFMR on all matching and retrieval tasks are declined. When the scale of detection box size is increased by 1.1 times, better performance can be obtained than the default settings. 

\textbf{Effect of voice anchored embedding learning.} By freezing the pre-trained voice embedding network, anchored embedding learning is conducted in the training process. As shown in Table \ref{table:bigtable2}, freezing the face embedding network reduces the performance, while freezing the voice embedding network improves the performance slightly. Human voices are related to some local features of human faces. Similar faces in traditional face recognition tasks do not necessarily have similar voices. Therefore, voice anchored embedding learning outperforms face anchored embedding learning. Training efficiency is improved greatly by freezing the voice network.

\begin{figure}[!tbp]
\centering
\includegraphics[width=1\linewidth]{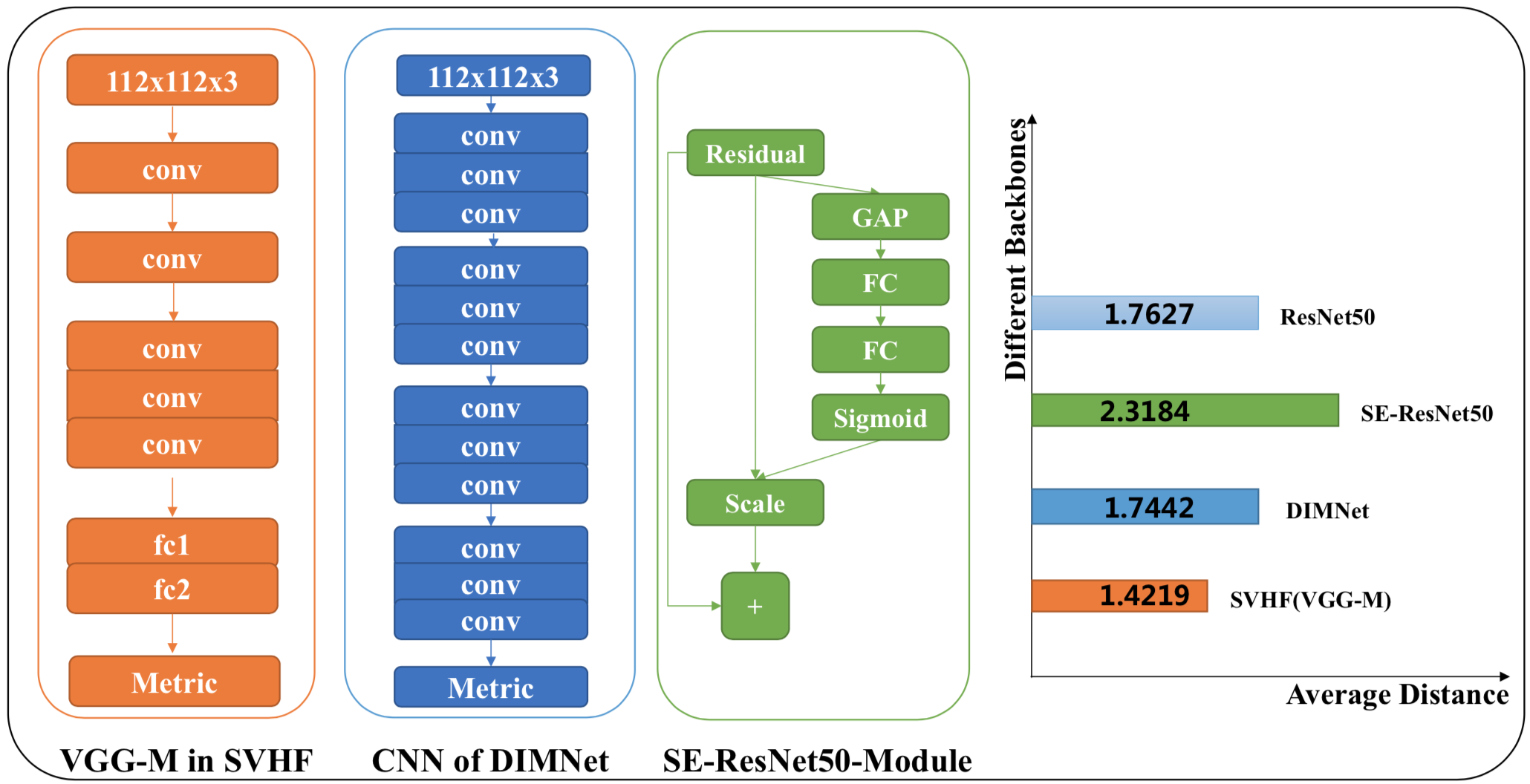}
\caption{Illustration of average distances between positives and negatives when different CNN backbones are used.}
\label{fig:cnnbackbone}
\end{figure}

\textbf{Effect of different network structures.} Generally, the best performed network structures on speaker identification and face recognition also perform well on the voice-face cross modal learning problem. As shown in Table \ref{table:bigtable2}, much deeper structures of CNN such as SE-ResNet50 and the structure used in DIMNet  outperform traditional shallow structures such as VGG-M. SE-ResNet50 with the squeeze-and-excitation module \cite{Jie2017Squeeze} outperforms the original ResNet50 structure used in the default settings. The mean distances between the embeddings of positive faces and negative faces learned by the four networks on 1:2 matching task are also illustrated in Figure \ref{fig:cnnbackbone}. Compared to other structures, positives and negatives learnt by SE-ResNet50 are more distant.

\textbf{Pre-training for both face network and voice network are needed.} 
As shown in Table \ref{table:bigtable2}, though the effect of pre-training with a larger data set MS1M \cite{guo2016ms} is not obvious, model performance is greatly reduced without pre-training (In this case, the voice network is not frozen).

\textbf{Effect of metric learning.} In the default configuration, the size of the metric space is 128. The model performance descends when the scale is set to 1, which indicates that it is necessary to properly increase the size of metric space.

\begin{table}[tp]
\centering
\scalebox{0.5}{
\begin{tabular}{c|c|c|c|c|c|c|c|c|c}
\hline
Config & MTCNN & Crop Audio & Network(Face) & Metric & Scale & Face Frozen & Voice Frozen\\
\hline

Default & 1.0 & Whole & LResNet50 & L2 & 128 & N & Y\\
\hline
  \end{tabular}
}
\caption{VFMR1 default config.}
\label{table:default config}
\end{table}

\begin{table}[tp]
\centering
\scalebox{0.62}{
\begin{tabular}{c|c|c|c|c|c|c|c|c}
\hline
Config & \multicolumn{2}{c|}{Details} & 1:2 & 1:3 & 1:4 & Male & Female & mAP\\
\hline

Default & \multicolumn{2}{c|}{Default Training Config} & \textbf{84.48} & \textbf{73.50} & \textbf{64.43} & \textbf{69.85} & \textbf{71.04} & \textbf{10.67}\\

\hline
\multirow{4}{*}{\makecell[c]{Prepro-\\cessing}} & MTCNN & Crop Audio & \multicolumn{6}{c}{} \\
\cline{2-9}
& -   & Whole & 83.04 & 71.13 & 62.02 & 66.99 & 69.69 & 8.22\\
& 1.1 & Whole & \textbf{84.27} & \textbf{73.28} & \textbf{65.55} & \textbf{70.03} & \textbf{71.65} & \textbf{10.42} \\
& 1.0 & 2.5s & 83.11 & 70.49 & 61.41 & 64.88 & 69.45 & 9.09\\

\hline
\multirow{3}{*}{\makecell[c]{CNN\\Frozen}} & Face Frozen & Voice Frozen & \multicolumn{6}{c}{} \\
\cline{2-9}
& N & N & \textbf{84.43} & \textbf{73.56} & \textbf{64.59} & \textbf{68.27} & \textbf{68.85} & \textbf{9.77}  \\
& Y & N & 81.68 & 68.52 & 59.43 & 64.87 & 66.76 & 8.33  \\

\hline
\multirow{4}{*}{Network} & \multicolumn{2}{c|}{Network} & \multicolumn{6}{c}{}\\
\cline{2-9}
& \multicolumn{2}{c|}{SE-ResNet50} & \textbf{84.00} & \textbf{72.62} & \textbf{63.70} & 69.22 & \textbf{69.49} & 9.40\\
& \multicolumn{2}{c|}{VGG-M} & 81.30 & 68.95 & 59.13 & 64.90 & 64.84 & 8.06 \\
& \multicolumn{2}{c|}{DIMNet} & 83.88 & 72.06 & 62.93 & \textbf{69.67} & 68.98 & \textbf{10.05} \\

\hline
\multirow{4}{*}{Pre-train} & Pre-Face & Pre-Voice & \multicolumn{6}{c}{}  \\
\cline{2-9}
& MS1M & VOX2 & \textbf{84.13} & \textbf{71.71} & \textbf{64.74} & \textbf{63.81}& \textbf{74.70} & \textbf{10.47} \\
& None & VOX2 & 73.29 & 58.20 & 48.72 & 54.26 & 58.90 & 5.56 \\
& None & None & 70.64 & 54.11 & 44.18 & 52.21 & 51.47 & 3.91 \\

\hline
\multirow{4}{*}{Metric} & Metric & Scale & \multicolumn{6}{c}{}  \\
\cline{2-9}
& None & 128 & 82.19 & 69.29 & 60.60 & 69.38 & 64.97 & 8.44 \\
& L2 & 1 & 81.71 & 69.09 & 59.61 & 62.62 & 67.49 & 7.58 \\
& L2 & 512 & \textbf{84.27} & \textbf{72.48} & \textbf{64.00} & \textbf{69.43} & \textbf{71.77} & \textbf{10.02} \\

\hline
  \end{tabular}
}

\caption{Comparisons of different submodule training settings for VFMR1 on matching and retrieval tasks.}
\label{table:bigtable2}
\end{table}

\section{Acknowledgement}

This work was supported by National Natural Science Foundation of China Grant Nos. U1711262 and 61472428.

\section{Conclusion}

A benchmark is established for voice-face matching and retrieval. Contributions of this paper involves the whole process of voice-face cross-modal learning, including:

A dataset collection tool for type-specific videos is developed. By taking advantage of the widely available MOOC videos, large scale dataset can be constructed efficiently with minimum labeling effort using this tool. A voice-face dataset of 500 Chinese speakers is constructed. By fine-tuning the pre-trained model on 1:2 matching task, 79\% of accuracy can be obtained on Chinese speakers. Much better results can be achieved after further expanding the data scale with the data collection tool.

A state-of-the-art voice-face matching and retrieval method is proposed, which is tested on large scale data set with high test confidence. On 1:2 matching and retrieval tasks, VFMR1 achieves the accuracy of 84.48\% and the mAP of 11\%. Compared to the best results published so far, the improvement for mAP is 7 percent. For seen and heard test on 1:2 matching and retrieval task, 95\% of accuracy and 38 percent mAP can be obtained when small scale training and test set are used, which shows the potential to be applied in practical scenarios with pre-registered persons.

{\small
\bibliographystyle{ieee_fullname}
\bibliography{egbib}
}

\end{document}